\documentclass[letterpaper]{article}
\usepackage{aaai}
\usepackage{times}
\usepackage{helvet}
\usepackage{courier}
\usepackage{algorithm,algorithmicx,algpseudocode}
\usepackage{csvsimple,longtable}
\usepackage{booktabs}
\usepackage{array}
\usepackage{datatool}
\usepackage{graphicx}
\usepackage{subcaption}
\usepackage{datatool}
\usepackage{amssymb}
\usepackage{xstring}
\usepackage{balance}
\usepackage{dblfloatfix}
\begin{document}
%
\title{Implementing Fair Regression In The Real World}

\author{Boris Ruf,
Marcin Detyniecki\\
AXA, AI Research\\
Paris, France\\
\{boris.ruf, marcin.detyniecki\}@axa.com
}

\nocopyright

\maketitle

\begin{abstract}
\begin{quote}
Most fair regression algorithms mitigate bias towards sensitive sub populations and therefore improve fairness at group level. In this paper, we investigate the impact of such implementation of fair regression on the individual. More precisely, we assess the evolution of continuous predictions from an unconstrained to a fair algorithm by comparing results from baseline algorithms with fair regression algorithms for the same data points. Based on our findings, we propose a set of post-processing algorithms to improve the utility of the existing fair regression approaches. \end{quote}
\end{abstract}

\noindent The potential risk of machine learning algorithms to unintentionally embed and reproduce bias and therefore discriminating various sub populations in high-stakes decision-making applications has given rise to the new research field of fair machine learning~\cite{Kamiran2009,Corbett-Davies2018,barocas-hardt-narayanan}. Plenty of quantitative measures of fairness have been proposed~\cite{Dwork2011,Hardt2016,Chouldechova2017,Berk2018} which opened up the way for three types of algorithms that seek to satisfy them: First, the pre-processing approach which modifies the data representation prior to using classical algorithms~\cite{Kamiran2012,Zemel2013}. Second, the in-processing approach which intervenes during the learning phase by adding a fairness constraint to the optimization objective~\cite{Kamishima2012,Zafar2017,Zhang2018}. Third, the post-processing approach which adjusts the outputs of classical algorithms~\cite{Hardt2016,Pleiss2017}.

So far, most presented solutions address classification problems where the decision space is binary. In practice, this covers use cases where the prediction is categorical, such as accepting or rejecting loan, school admission or job applications. However, many real-world problems describe regression tasks where the decision space is continuous, for example algorithms for pricing insurance premiums or establishing credit card limits. Further, stakeholders often prefer scores or ranks even for categorical tasks because they allow more room for human intuition and interpretation~\cite{Veale2018}. It is therefore a welcomed step that first approaches in the direction of fair regression have been proposed lately~\cite{Komiyama2018,Speicher2018,Agarwal2019,grari2019fairnessaware}.

Just as in most fair classification methods, fair regression approaches seek to satisfy statistical fairness at group level. Due to the real-valued character of regression predictions, the fairness metrics optimize for the average outcome. 
In this paper, we investigate the impact of this objective at an individual level\footnote{To avoid misunderstandings with regard to the term ``individual'', we would like to stress that we assess the impact of group fairness measures on individuals in this work; it is not about individual fairness measures.}. We compare the different outcomes of unconstrained and fair methods for the same set of data. While the fair methods attain global fairness, we observe significant variations for some of the data points. In a business context where an unconstrained real-world application were to be replaced with a fairer one, such extreme discrepancies would not be viable because individuals who were substantially negatively impacted would probably not accept the change and switch to a competitor. Based on our findings, we therefore propose algorithmic post-processing procedures to adjust for unwanted, extreme discrepancies between unconstrained and fair methods in order to enable a smooth transition from an ``unfair'' to a fairer model.

 The main contributions of this paper are:
 
 \begin{itemize}
 \item We empirically examine the evolution of fair regression outputs compared to unconstrained predictors and demonstrate that some variations on the individual level may be unacceptable in practice. To the best of our knowledge we offer the first investigation of this kind;
 \item We propose a range of post-processing algorithms to mitigate this effect and therefore provide mechanisms to implement fair regression in practice. In particular, we approach two major real-world aspects: seamless market adoption which could be realized by aiming at non-positive outcome evolution, and the economical impact on the company which could be minimized by targeting budget neutrality;
 \item We study our proposals on three different real data sets and clarify the trade-offs between both competing aspects.

 \end{itemize}

The remainder of this paper proceeds as follows. First, we recap the currently available solutions for fair regression problems in the following section. Second, we define our notion of fairness and clarify the fair regression task in Section~\ref{sec:fair_regression}. Next, we describe our evaluation methodology of fair regression predictions on individual level in Section~\ref{sec:evaluation}. Finally, we propose in Section~\ref{sec:mitigation_proposals} mitigation approaches for two realistic scenarios: non-positive evolution and budget neutrality. We conclude with a discussion of the benefits and the drawbacks of our suggestions.

\section{Related Work}
\label{sec:related_work}

Some prior work with respect to fair predictors of real-valued targets has been published. Regularization approaches which seek to fit probabilistic models that satisfy statistical independence were introduced~\cite{Kamishima2012,Fukuchi2015}. Propensity modeling was proposed to control the biasing effect of a sensitive attribute in linear regression models~\cite{Calders2013}. A convex framework for fair linear and logistic regression problems was considered~\cite{Berk2017}. Fair regression methods built on the Hilbert-Schmidt Independence Criteria have been introduced~\cite{Perez-Suay2017}. Algorithmic unfairness in regression problems was studied using inequality indices~\cite{Speicher2018}. A non-convex optimization approach for fair regression was presented ~\cite{Komiyama2018}. Efficient fair regression algorithms which achieve full statistical independence for arbitrary model classes were proposed~\cite{Agarwal2019}.

\section{Fairness and Fair Regressions}
\label{sec:fair_regression}
Throughout this paper we assume the problem of predicting a real-valued target while guaranteeing fairness towards a sensitive group. More precisely, we consider a regression task with a source distribution over $(X,A,Y)$, where $X$ are the available features, $A$ is the sensitive attribute which defines the membership to the protected group and $Y \in [0,1]$ is the continuous output. 

\emph{Fairness Definition.}
We consider demographic (or statistical) parity as quantitative definition of fairness. This criterion requires the prediction to be statistically independent from any sensitive group membership and has been widely applied for fair classification~\cite{Dwork2011}. For regression problems, we adopt the definition from \cite{Agarwal2019}.

\newtheorem{dfn}{Definition}
\begin{dfn}
(Demographic parity–DP). A fair predictor $f$ satisfies demographic parity under a distribution over $(X,A,Y)$ if $f(X)$ is independent of the sensitive attribute $A$. Since $f(X) \in$ [0,1] as per definition, this corresponds to $\mathbb{P}[f(X)\geq z|A=A]=\mathbb{P}[f(X)\geq z]$ for all $a\in A$ and $z \in [0,1]$.
\end{dfn}

The real-valued random variable $f(X)$ is fully characterized by its cumulative distribution function (CDF). We compute the DP disparity as the difference between the CDFs which is measured in the $l_\infty$ norm, analogous to the Kolmogorov-Smirnov (KS) statistic~\cite{Lehmann2005}.  

\section{The Impact of Fairness} 
\label{sec:evaluation}
We investigate the evolution of regression outputs by comparing unconstrained baseline algorithms with fair algorithms. As baseline algorithms we use a simple logistic regression (LR) learner and a XGBoost classifier for logistic loss tasks. For least square tasks, we use an ordinary least squares (OLS) learner, a XGBoost regressor and a standard Random Forest (RF) regressor. For the fair algorithms we follow the approach of \cite{Agarwal2019} which discretizes the real-valued prediction space and then reduces the fair regression problem to cost-sensitive classification. A slack parameter $\varepsilon$ allows to tune the trade-offs between fairness and accuracy.\footnote{We leave it at this very short description and refer the interested reader to \cite{Agarwal2019} for more details since our focus is not on any specific algorithm but rather on studying the general impact of bias mitigation for the individual.}

In the absence of standard benchmarking data sets for fair regression, we follow the same practice as \cite{Agarwal2019} and use popular data sets for fair classification problems but transform them into regression problems by modifying the prediction task. Precisely, we use the following data sets as described below:

\emph{Adult:} The adult income data set contains 48,842 instances~\cite{Dua:2019}. The sensitive attribute is a boolean value which represents the gender. The task is to predict via logistic loss minimization the probability that a person has a salary of more than \$50k per year.

\emph{Communities \& crime:} The data set contains 1,994 instances of socio-economic, law enforcement, and crime data about communities in the United States~\cite{Redmond2002ADS}. The sensitive attribute is a boolean value which states whether the majority population of the community is white. The task is to predict via square loss minimization the number of violent crimes per 100,000 population, normalized to [0, 1].

\emph{Law school:} The LSAC National Longitudinal Bar Passage Study contains 20,649 instances~\cite{Wightman1998LSACNL}. The sensitive attribute is a boolean value which states whether the person is white or not. The task is to predict via square loss minimization a student's GPA, normalized to [0, 1]. 

For training the models, we randomly split the data in half:
50\% for training and 50\% for testing. For comparison of the baseline and the fair outputs we use 1,000 randomly selected data points and apply them to both trained models. We denote the output of the baseline model as $B$ and the output of the fair model as $F$. To assess the individual variations of the data points, we compute the distribution of differences $D$ by subtracting $B$ from $F$.

For the full list of results for each data set, please see Table~\ref{tab:results_adult}, Table~\ref{tab:results_communities} and Table~\ref{tab:results_law_school} at the end of this document. Each line of the table describes one experiment. The first group of columns provides the performance of the ``unfair'' baseline algorithm in terms of accuracy (``Loss STD'') and fairness (``DP disp''). The second group provides the same performance metrics for the fair algorithm we used for comparison. It also contains the heuristically chosen slack parameter $\varepsilon$ which adjusts the trade-off between fairness and accuracy. In the last group of columns, we provide the comparative analysis: We record the most extreme variations observed in this experiment for a given data point (``Max increase'' and ``Max decrease''), as well as the average difference when comparing the baseline outputs with the ones from the fair algorithm. For each table, the maximum values are highlighted in bold. For all three data sets we observe minor, mostly negative deviations for the average difference. The most extreme individual variations turn out to be very high though, for example for the Adult data set in Table~\ref{tab:results_adult} the maximum increases for some experiments range above 100\%.

\begin{figure}[h]
  \centering
  \includegraphics[width=0.4\textwidth]{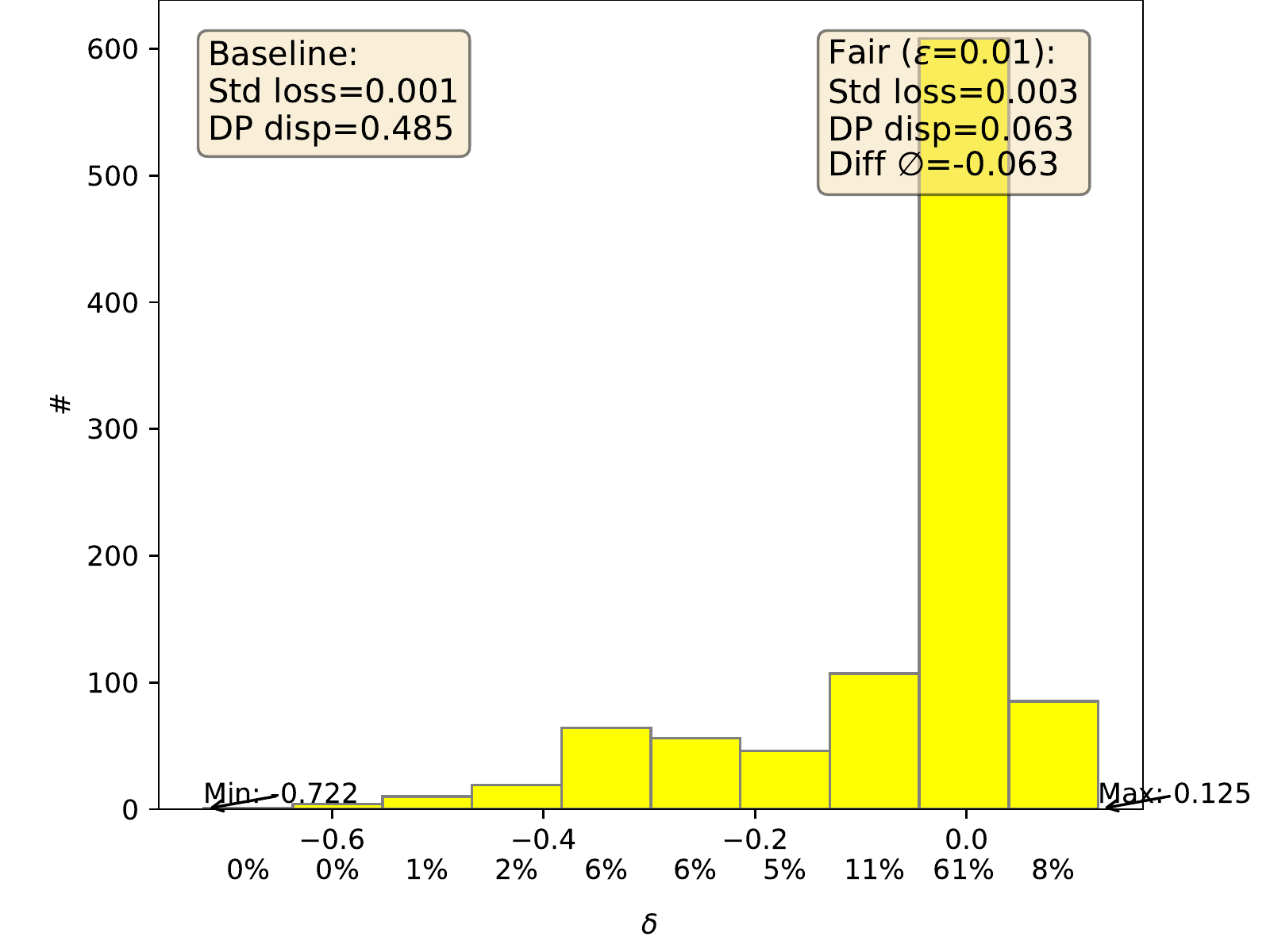}
  \caption{Histogram describing the distribution of differences $D$, comparing the baseline (OLS) to fair output on the individual level for the Communities data set}
  \label{fig:comparison}
\end{figure}

Throughout this document, we illustrate our findings by plotting the distribution of differences $D$ as histograms. The x-axis describes the individual variations $\delta$ when comparing baseline with fair outputs, the y-axis represents the number of occurrences of the variations within a bin. Figure~\ref{fig:comparison} shows the results of one of our experiments for the Communities data set. We use an ordinary least squares (OLS) learner as unconstrained baseline. The standard loss for this algorithm is $0.001$ and the demographic parity (DP) disparity–in this case representing the disparity between men and women–is $0.485$. As fair learner we obtain a least squares (LS) regression model from the \textit{supervised learning oracle}~\cite{Agarwal2019} using the slack parameter $\varepsilon=0.01$. The standard loss is $0.003$ and therefore higher than for the baseline algorithm, but the DP disparity is significantly lower ($0.063$). The average difference of the fair output with respect to the baseline is $-0.063$. The most extreme differences are a decrease of $0.722$ on the lowest end and an increase of $0.125$ on the highest end. In the context of the data set, this means that for one community the prediction of violent crimes decreases by about 72 percent points while for another community it increases by about 12 percent points compared to the original, ``unfair'' model.

When we look at these results in a business context, we find that an implementation in practice is not realistic. Applied on pricing models, for example, the extreme increases between the conventional baseline model and the fair model would not be acceptable for the concerned individuals and likely lead to customer churn in a competitive market. Further, if the average difference is negative, this corresponds to a loss of revenue compared to the baseline algorithm. This ``cost of fairness'' would need to be managed in any application with economically viable objective.

\section{Post-processing Algorithms}
\label{sec:mitigation_proposals}
In the previous section, we demonstrated the impact of fair regression algorithms on the individual level and explain how such shortcomings pose substantial practical obstacles when trying to upgrade existing machine learning applications to fair machine learning applications in real-world applications. In order to counter those findings, we propose two different, rather simple post-processing approaches to provide actionable methods which mitigate the unwanted effects and help make the final outputs more suitable in practice. For each operation, we briefly describe their real-world consequences in a sample insurance pricing scenario.

\subsection{Non-positive Evolution}
\label{sec:non-positive_evolution}

The first family of post-processing algorithms produces fair outcomes which are always lower or unchanged compared to the baseline outcomes. In a pricing scenario, such a ``non-positive evolution'' would correspond to unchanged or discounted rates only – something that should be easily acceptable by the customer base.

\begin{figure*}[h]
  \centering
  \begin{subfigure}{.4\textwidth}
    \includegraphics[width=1.0\textwidth]{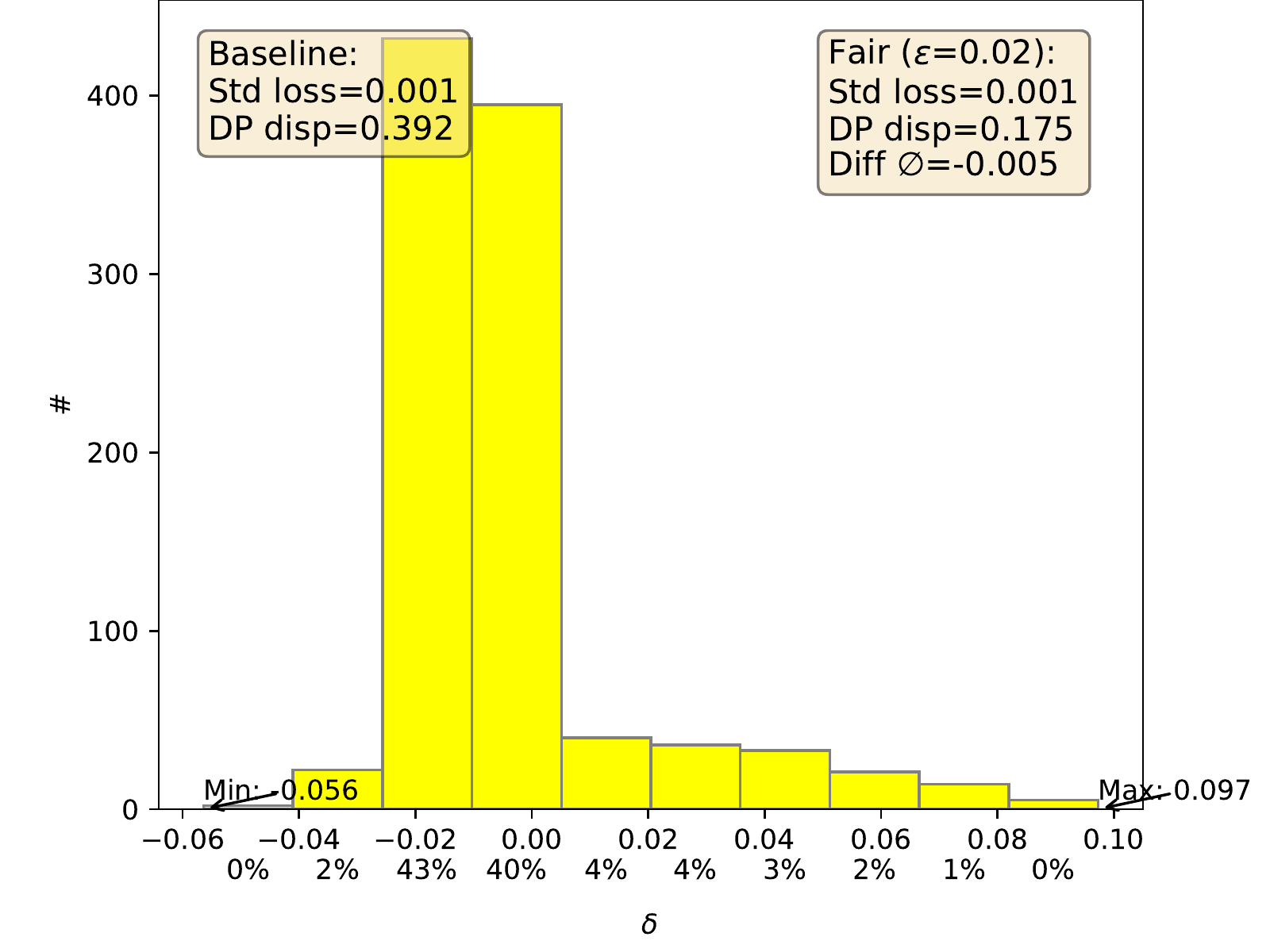}
    \vspace{-0.6cm}
    \caption{Without post-processing}
    \label{fig:non_positive_without}
  \end{subfigure}
  \hspace{0.6cm}
  \begin{subfigure}{.4\textwidth}
    \includegraphics[width=1.0\textwidth]{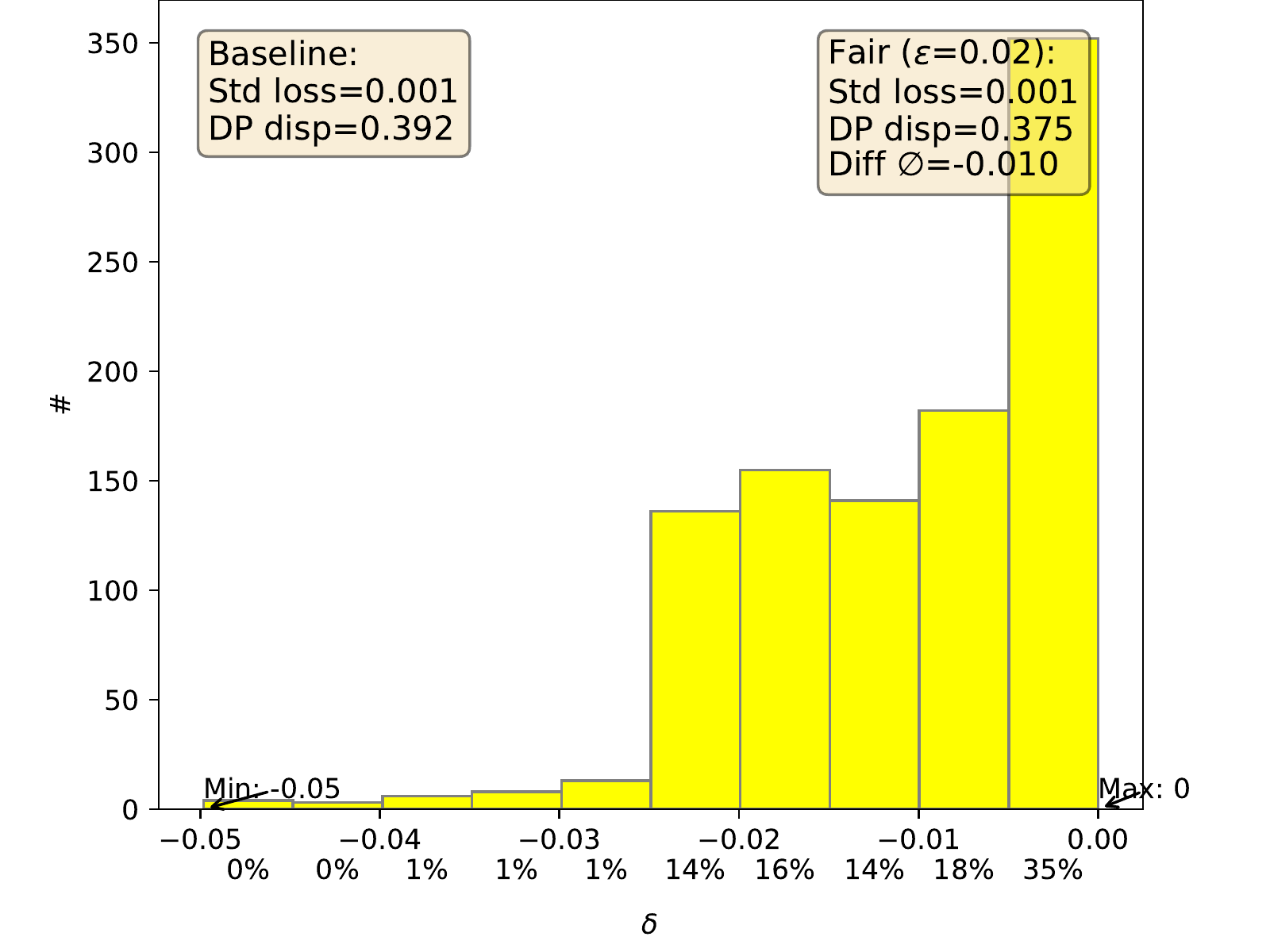}
    \vspace{-0.6cm}
    \caption{Capped at $\theta=0$}
    \label{fig:non_positive_capped}
  \end{subfigure}
  \par\bigskip
  \begin{subfigure}{.4\textwidth}
    \includegraphics[width=1.0\textwidth]{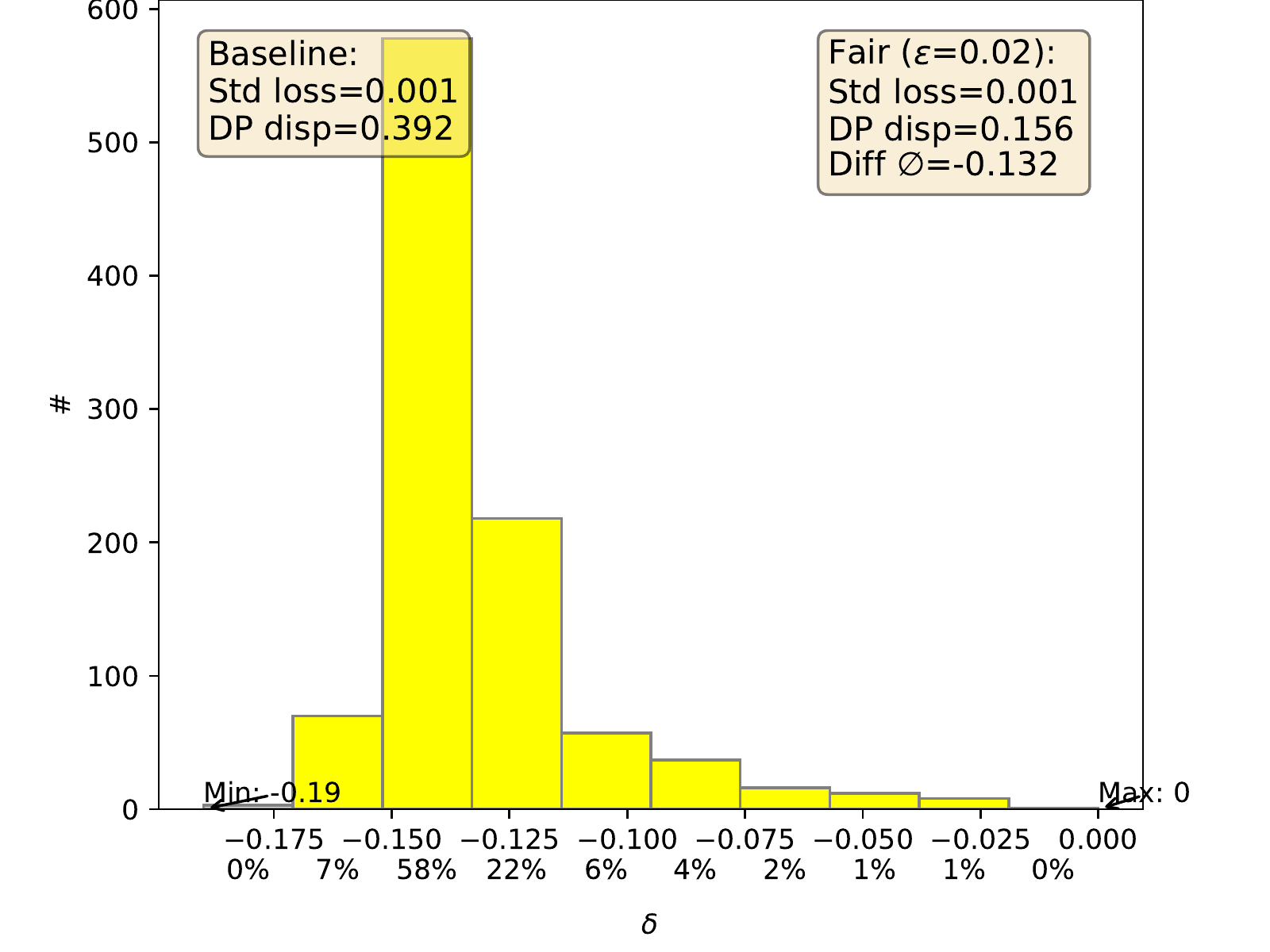}
    \vspace{-0.6cm}
    \caption{Translated by $-arg max(\delta)$}
    \label{fig:non_positive_translated}
  \end{subfigure}
  \hspace{0.6cm}
  \begin{subfigure}{.4\textwidth}
    \includegraphics[width=1.0\textwidth]{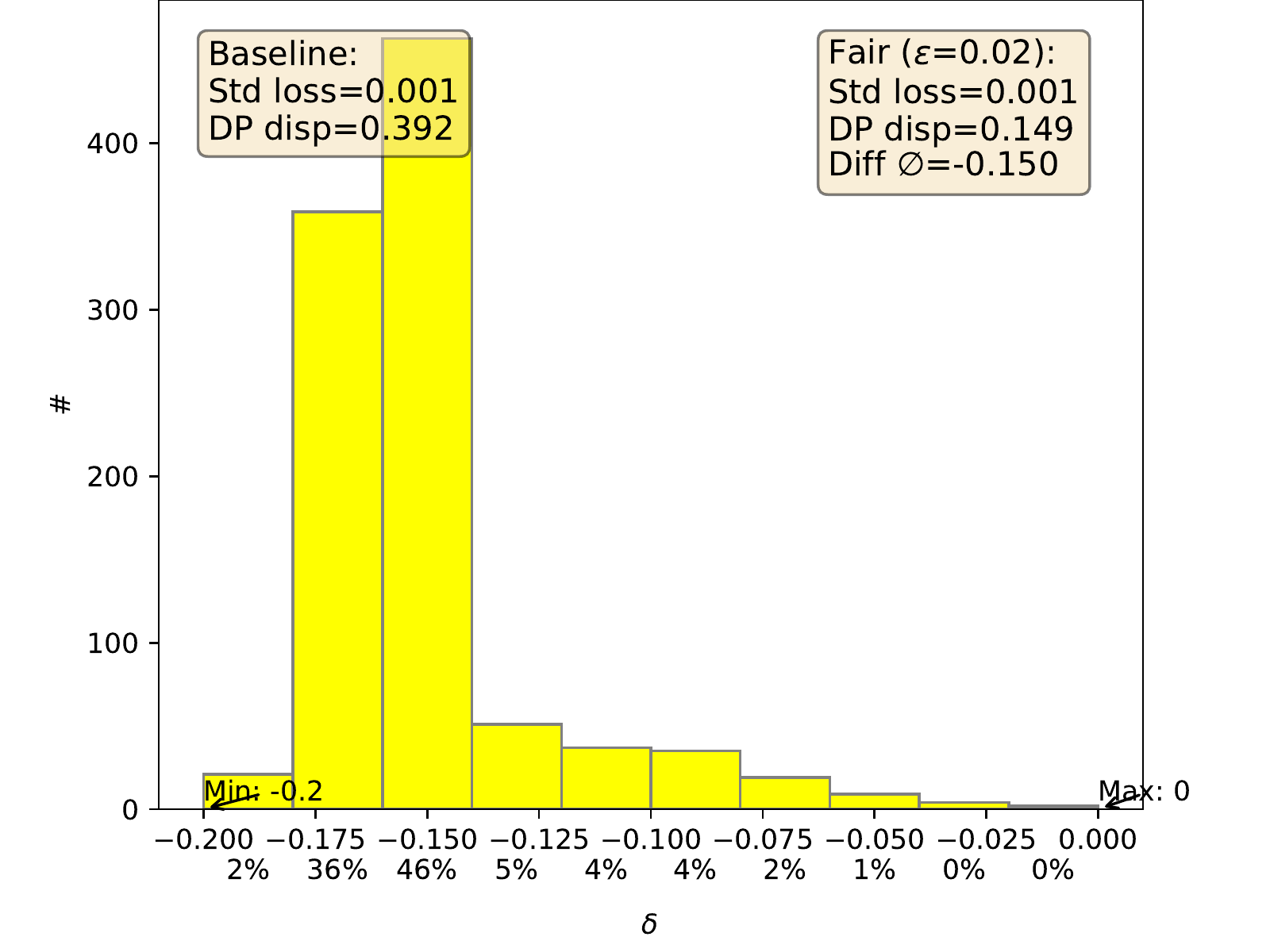}
    \vspace{-0.6cm}
    \caption{Normalized to $[-0.2, 0]$}
    \label{fig:non_positive_normalized}
  \end{subfigure}

  \caption{Experimental results with the objective of non-positive evolution for the Law School data set}
  \label{fig:non_positive}
\end{figure*}

In the following, we consider several steps in order to achieve this objective. We provide pseudo codes which describe the approaches and empirical evaluate the results on the Law School data set. As unconstrained baseline we use an ordinary least squares (OLS) learner. As fair learner we use a least squares (LS) regression model which was returned from the \textit{supervised learning oracle} for $\varepsilon=0.02$. See Figure~\ref{fig:non_positive} for the results of each step. Figure~\ref{fig:non_positive_without} is the starting point, illustrating the distribution of differences prior to any post-processing of the fair output.

First, we propose to cap any fair outcome if it exceeds the baseline outcome. Concretely, any positive value in the distribution of differences $D$ is set to 0 and the resulting distribution is added to the baseline distribution. Generic pseudo code for this operation has been outlined in Algorithm~\ref{alg:algorithm_cap}. The algorithm uses a variable threshold $\theta$ which defines the maximum accepted increase in individual output. The approach described in this paragraph corresponds to $\theta=0$. An empirical evaluation can be found in Figure~\ref{fig:non_positive_capped}. As expected, we notice that this step suppresses any positive variation: The maximum increase of fair predictions compared to the baseline is now 0, while the maximum decrease remains unchanged. However, we also observe that all predictions which had originally produced a positive variation fall into the same rightmost bin in this scenario, and their previous rank order is gone. 

Put in an insurance pricing context, customers who were to pay more without post-processing continue to pay the same price. Customers who profit from reduced rates continue to do so. However, the sum of collected primes and as such the total revenue for the company sinks, too, which poses a problem for implementation. A final observation is that if price increases are not ruled out in principle, the threshold parameter $\theta$ serves as useful means to restrict them.

\begin{algorithm}[h]
\begin{algorithmic}
   \State{\bfseries Inputs:} Baseline outputs $B_{[n]}=\{b_i\}_{i=1}^{N}$, Fair outputs $F_{[n]}=\{f_i\}_{i=1}^{N}$, Threshold $\theta$
   \State{}
   \State{Let $Y_{[n]} := F_{[n]}$} \Comment{Initialize output with fair predictions}
   \For{i=1, ..., N}
     \If{$f_i > b_i + \theta$} \Comment{Cap if fair value is above threshold}
       \State{Let $y_i := b_i + \theta$}
     \EndIf
   \EndFor
   \State{\bfseries Return} $Y_{[n]}$
\end{algorithmic}
\caption{Limit the increase of the fair outputs to a given maximum threshold $\theta$.}
\label{alg:algorithm_cap}
\end{algorithm}

Next, we propose a different approach. Instead of capping any positive variation, we  reduce all fair outputs by the highest recorded positive difference compared to the baseline. As a result, the distribution of differences $D$ moves in negative direction by the magnitude of its maximum positive variation. Pseudo code describes this method in  Algorithm~\ref{alg:algorithm_translate_non_positive}, empirical results can be studied in Figure~\ref{fig:non_positive_translated}. We notice that the result is a non-positive evolution on the individual level, because the outcome with the maximum increase in the fair prediction is unchanged in the final output. Since the distances within the fair distribution are preserved, any other output gets decreased. Another notable property of this operation is that the DP disparity remains unchanged. However, the average difference and the minimum both decrease compared to the original distribution.

In the pricing scenario, again no customer sees her prime increase. For some customers, discounts turn out to be much higher compared to before the post-processing. And the average difference decreases significantly, meaning this implementation of fairness comes at an increased cost for the company.


\begin{algorithm}[h]
\begin{algorithmic}
   \State{\bfseries Inputs:} Baseline outputs $B_{[n]}=\{b_i\}_{i=1}^{N}$, Fair outputs $F_{[n]}=\{f_i\}_{i=1}^{N}$
   \State{}
   \State{Let $\Delta := argmax(F_{[n]}-B_{[n]})$} \Comment{Identify the maximum increase}
   \State{Let $Y_{[n]} := F_{[n]} - \Delta$} \Comment{Shift the outputs by $\Delta$}
   \State{\bfseries Return} $Y_{[n]}$
\end{algorithmic}
\caption{Achieve non-positive evolution by translation of the distribution of differences.}
\label{alg:algorithm_translate_non_positive}
\end{algorithm}

Finally, we propose to first normalize the distribution of differences $D$ to an arbitrary range limited by minimum $a$ and maximum $b$:

\begin{eqnarray}
D := a + \frac{(D - min(D))(b-a)}{max(D)-min(D)}
\end{eqnarray}

Afterwards, we subtract the highest positive value of the resulting distribution in order to obtain non-positive values as described in the previous step. Generic pseudo code with variable min and max parameters $a$ and $b$ for the desired range has been outlined in Algorithm~\ref{alg:algorithm_translate_norm_non_positive}. Results of this approach are shown in Figure~\ref{fig:non_positive_normalized}, where the min parameter has been heuristically set to $a=-0.2$, and the max parameter ensures with $b=0$ the non-positive evolution. We notice that this approach provides some flexibility by allowing to limit the maximum decrease and the maximum increase to fixed values. It also preserves the original fair rank order.

Applied to the pricing scenario, this approach can enable the stakeholders to limit the maximum discount while preventing any price increase. But it does not guarantee equal overall revenue compared to the baseline model. 

\begin{algorithm}[h]
\begin{algorithmic}
   \State{\bfseries Inputs:} Baseline outputs $B_{[n]}=\{b_i\}_{i=1}^{N}$, Fair outputs $F_{[n]}=\{f_i\}_{i=1}^{N}$, Min-max values $a$ and $b$
   \State{}
   \State{Let $\Delta_{[n]} := F_{[n]}-B_{[n]}$} \Comment{Compute the deltas}
   \State{Let $Y_{[n]}:=B_{[n]}-a+(\Delta_{[n]}-argmin(\Delta_{[n]}))$\Comment{Normalize}\par $/(arg min(\Delta_{[n]})-arg max(\Delta_{[n]}))*b$} 
   \State{Let $\Gamma := argmax(Y_{[n]}-B_{[n]})$} \Comment{Identify the maximum increase}
   \State{$Y_{[n]} = Y_{[n]} - \Gamma$} \Comment{Shift the outputs by $\Gamma$}
   
   \State{\bfseries Return} $Y_{[n]}$
\end{algorithmic}
\caption{Achieve non-positive evolution by normalization and translation of the distribution of differences.}
\label{alg:algorithm_translate_norm_non_positive}
\end{algorithm}

\begin{figure*}[h]
  \centering
  \begin{subfigure}{.3\textwidth}
  \centering
    \includegraphics[width=1\textwidth]{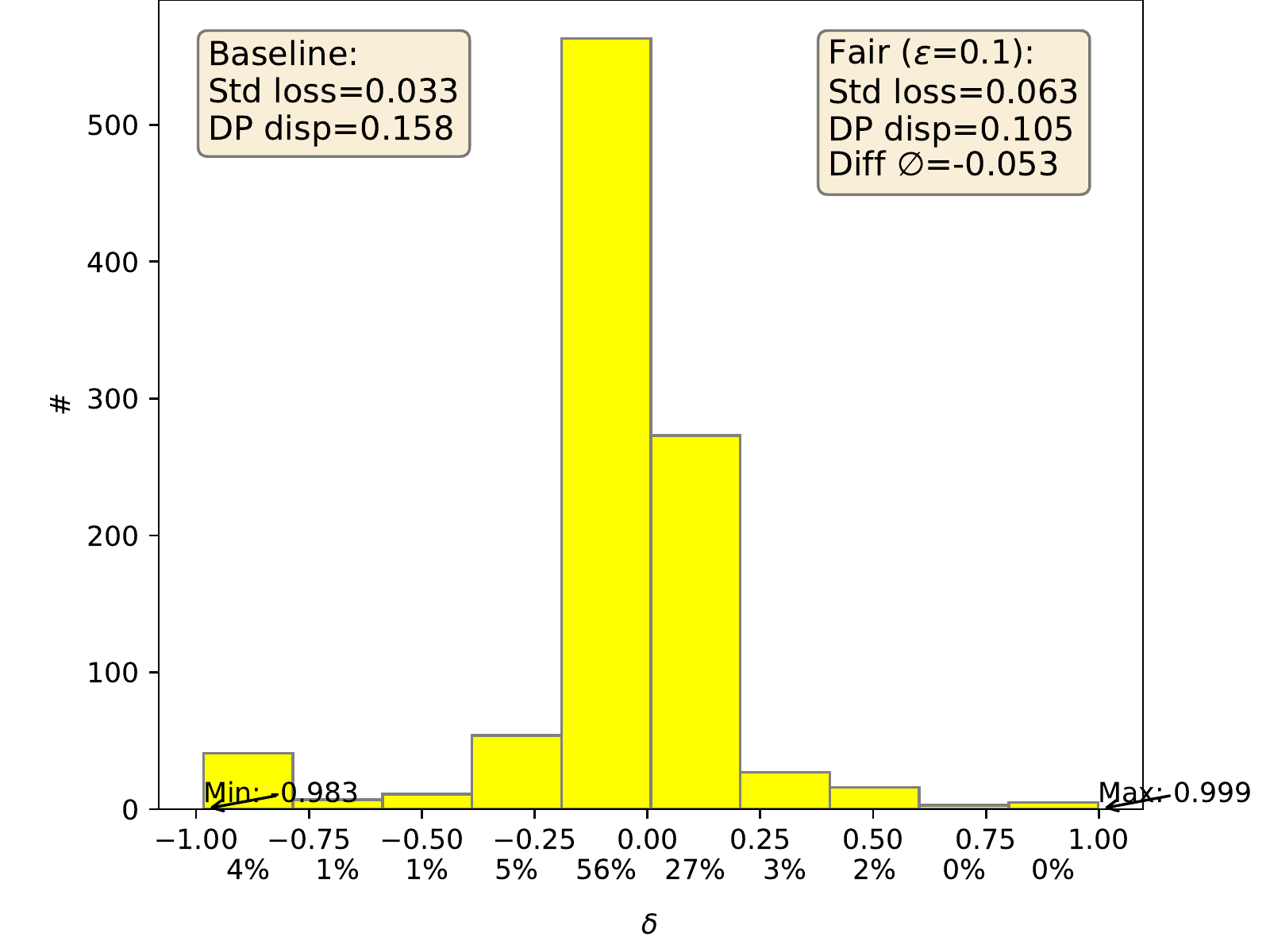}
    \caption{Without post-processing}
    \label{fig:budget_neutral_without}
  \end{subfigure}
  \hspace{0.6cm}
  \begin{subfigure}{.3\textwidth}
    \includegraphics[width=1\textwidth]{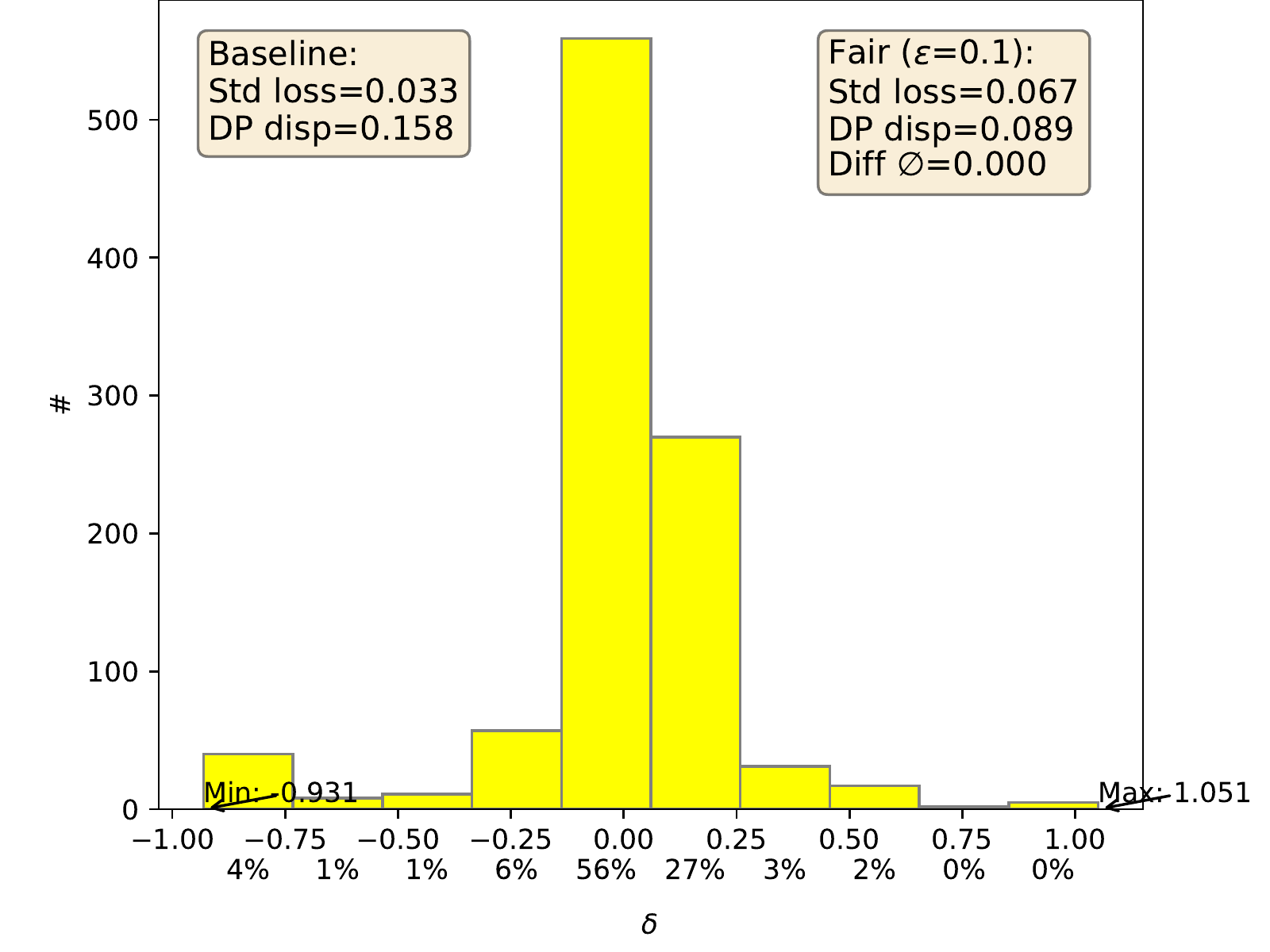}
    \caption{Translated by $-mean(\delta)$}
    \label{fig:budget_neutral_translated}
  \end{subfigure}
  \hspace{0.6cm}
  \begin{subfigure}{.3\textwidth}
    \includegraphics[width=1\textwidth]{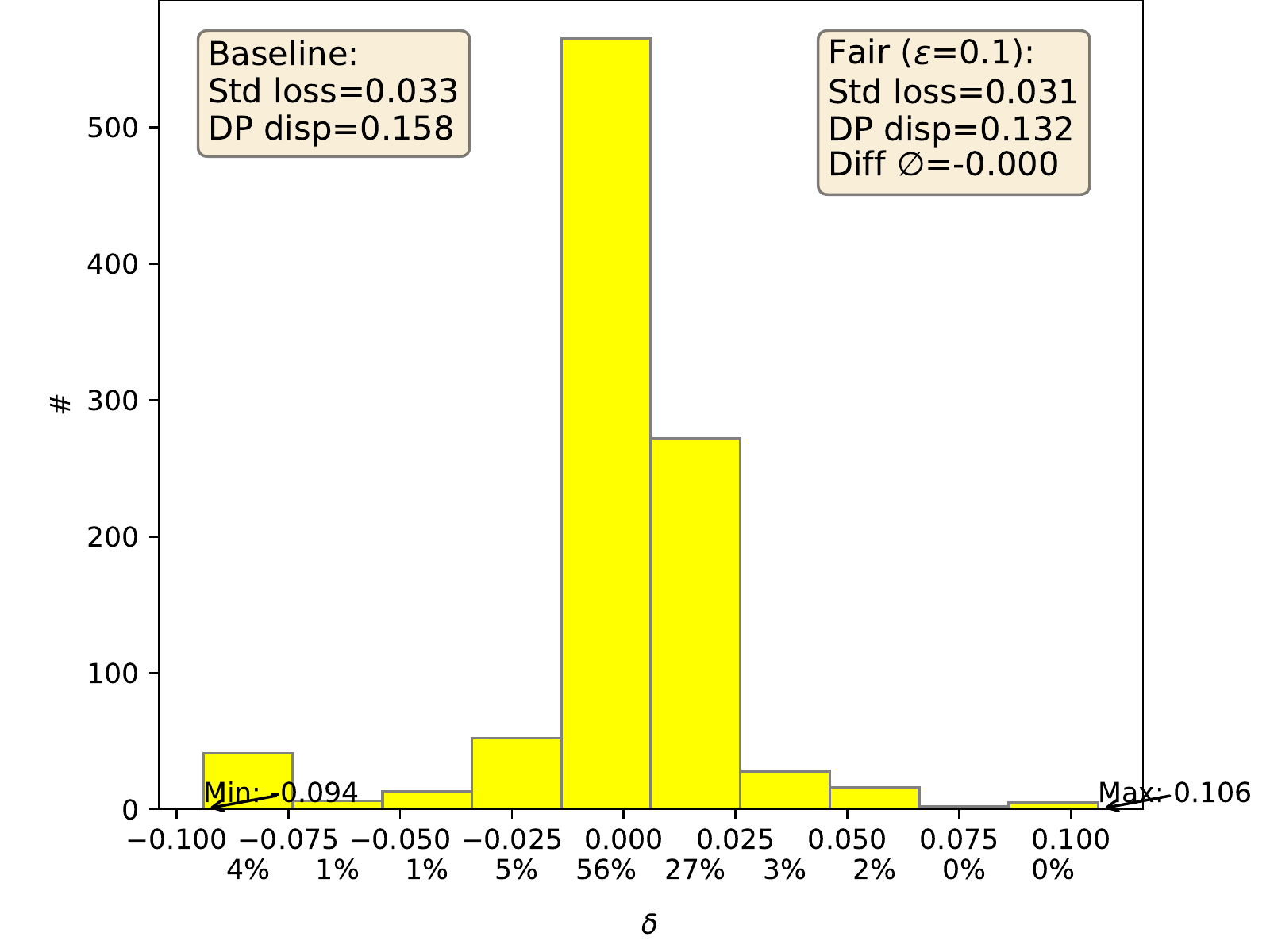}
  \caption{Normalized to $[-0.1, 0.1]$ and\\ translated by $-mean(\delta)$}
  \end{subfigure}
  \caption{Experimental results with the objective of budget neutrality for the Adult data set}
  \label{fig:budget_neutral}
\end{figure*}

\subsection{Budget Neutrality}
\label{sec:budget_neutrality}

The second type of post-processing algorithms we propose seeks to keep the mean of the fair distribution unchanged as compared to the baseline. In a pricing scenario, reducing the average value of the distribution of differences to 0 corresponds to constant overall revenue; the added fairness comes at no extra cost for the operator.

In this subsection, we describe two possible approaches in order to achieve this objective. We provide pseudo codes which explain both methods and empirical evaluate the results on the Adult data set. As unconstrained baseline we use a XGBoost classifier. As fair learner we use a logistic regression (LR) model which was returned from the \textit{supervised learning oracle} for $\varepsilon=0.1$. All results are illustrated in Figure~\ref{fig:budget_neutral}. As before, the first subfigure (Figure~\ref{fig:budget_neutral_without}) shows the distribution of differences without any post-processing.

First, we propose the straightforward approach of subtracting the average difference from the fair outputs. Looking at the distribution of differences, this corresponds to a shift in negative direction by the magnitude of the average variation. The result is an average difference of $0$ which describes a budget neutral operation in the given context. Pseudo code outlines this method in  Algorithm~\ref{alg:algorithm_translate_budget_neutrality}. Empirical results are displayed in Figure~\ref{fig:budget_neutral_translated}. We observe that this approach slightly decreases the accuracy, but the degree of fairness is improved.

In the pricing scenario, introducing fairness does not come at any additional cost for the company. However, for the customers the consequences are rather hefty: Some see their rates increased by a lot, while others get their prices greatly reduced.

\begin{algorithm}[h]
\begin{algorithmic}
   \State{\bfseries Inputs:} Baseline outputs $B_{[n]}=\{b_i\}_{i=1}^{N}$, Fair outputs $F_{[n]}=\{f_i\}_{i=1}^{N}$
   \State{}
   \State{Let $\Delta := mean(F_{[n]}-B_{[n]})$} \Comment{Identify the average diff}
   \State{Let $Y_{[n]} := F_{[n]} - \Delta$} \Comment{Shift the outputs by $\Delta$}
   \State{\bfseries Return} $Y_{[n]}$
\end{algorithmic}
\caption{Achieve budget neutrality by translation of the distribution of differences.}
\label{alg:algorithm_translate_budget_neutrality}
\end{algorithm}

As more elaborated approach we finally propose to combine min-max normalization, as introduced in subsection~\ref{sec:non-positive_evolution}, with the translation operation described above. When looking at the distribution of differences, this one gets first normalized to an arbitrary range, and afterwards the resulting distribution is translated by its average variation. Algorithm~\ref{alg:algorithm_translate_norm_budget_neutrality} provides pseudo code for a generic version of this method with a variable normalization range defined by the min and max parameters $a$ and $b$. Figure~\ref{fig:budget_neutral_translated} shows the experimental results for the heuristically chosen parameters $a=-0.1$ and $b=0.1$. We notice that the specified range was respected and the average difference is 0 as expected. The degree of fairness has slightly deteriorated, while accuracy improved.

In the insurance pricing context, the implementation barriers of this approach would be low due to the budget neutrality property. Also, the ability to control the maximum price increase as well as the lowest possible discount for the individual customers provide important instruments for introducing fair regression models in the real world.

\begin{algorithm}[h]
\begin{algorithmic}
   \State{\bfseries Inputs:} Baseline outputs $B_{[n]}=\{b_i\}_{i=1}^{N}$, Fair outputs $F_{[n]}=\{f_i\}_{i=1}^{N}$, Min-max values $a$ and $b$
   \State{}
   \State{Let $\Delta_{1[n]} := F_{[n]}-B_{[n]}$} \Comment{Compute the deltas}
   \State{Let $Y_{1[n]}:=B_{[n]}-a+(\Delta_{1[n]}-argmin(\Delta_{1[n]}))$\Comment{Normalize}\par $/(arg min(\Delta_{1[n]})-arg max(\Delta_{1[n]}))*b$} 
   \State{Let $\Delta_2 := mean(Y_{1[n]}-B_{[n]})$}
   \Comment{Identify the average diff}
   \State{Let $Y_{2[n]} := Y_{1[n]} - \Delta_2$} \Comment{Shift the outputs by $\Delta_2$}
   \State{\bfseries Return} $Y_{2[n]}$
\end{algorithmic}
\caption{Achieve budget neutrality by normalization and translation of the distribution of differences.}
\label{alg:algorithm_translate_norm_budget_neutrality}
\end{algorithm}

\section{Discussion and Conclusion}
In this paper we draw attention to the to date unnoticed problem of sometimes large prediction discrepancies at the individual level when replacing unconstrained regression algorithms with fair predictors that satisfy some sort of group fairness. We propose post-processing methods to mitigate this effect in order to tackle two real-world challenges: We first suggest to limit the outcome variation to non-positive values in order to enable a smooth market adoption. Second, we demonstrate mechanisms to achieve budget neutrality and therefore target the economical impact on a company when introducing fair regression methods. While translating the distribution has the beneficial property of preserving the fairness, this operation lowers the average difference and therefore comes at extra cost. On the other hand, only aiming for budget neutrality will either compromise the fairness or the error rate. Normalizing the distribution to a given range is a useful method to restrict the variation, but selecting the right bounds depends on the use case. After all, the most appropriate solution is a matter of context and will probably in most scenarios consist of a carefully chosen trade-off between both objectives. 


%
%
%


\newcommand{\HighlightCell}[2]{
    \IfStrEq{#1}{#2}{\textbf{#1}}{#1}
}

\begin{filecontents*}{data/adult.csv}
unfair learner,unfair loss std,unfair DP disp,fair eps,fair learner,state,fair loss std,fair DP disp,max individual increase,max individual decrease,average diff
LR,0.0685,0.1416,0.01,LR,0,0.0472,0.4713,1.011,-0.9253,0.0145
LR,0.0685,0.1416,0.02,LR,0,0.0487,0.4665,1.011,-0.9253,0.0054
LR,0.0685,0.1416,0.03,LR,0,0.0595,0.4583,1.0119,-0.9253,-0.0025
LR,0.0685,0.1416,0.04,LR,0,0.0508,0.4319,1.011,-0.9253,-0.0126
LR,0.0685,0.1416,0.06,LR,0,0.0632,0.3899,1.0119,-0.9253,-0.0278
LR,0.0685,0.1416,0.08,LR,0,0.0579,0.2681,1.011,-0.9366,-0.0429
LR,0.0685,0.1416,0.1,LR,0,0.0626,0.1071,1.011,-0.9616,-0.0572
LR,0.0685,0.1416,0.12,LR,0,0.0657,0.1122,1.011,-0.9616,-0.0659
LR,0.0685,0.1416,0.15,LR,0,0.066,0.1132,1.011,-0.9616,-0.0664
XGB Classifier,0.0331,0.1581,0.01,LR,0,0.0473,0.4705,0.9986,-0.9266,0.0179
XGB Classifier,0.0331,0.1581,0.02,LR,0,0.0482,0.4603,0.9986,-0.9472,0.0107
XGB Classifier,0.0331,0.1581,0.03,LR,0,0.0594,0.4625,0.9986,-0.9694,0.003
XGB Classifier,0.0331,0.1581,0.04,LR,0,0.051,0.4319,0.9986,-0.9542,-0.0083
XGB Classifier,0.0331,0.1581,0.06,LR,0,0.0632,0.3865,0.9986,-0.9766,-0.0239
XGB Classifier,0.0331,0.1581,0.08,LR,0,0.0581,0.2615,0.9986,-0.9819,-0.04
XGB Classifier,0.0331,0.1581,0.1,LR,0,0.0626,0.1049,0.9986,-0.9834,-0.0527
XGB Classifier,0.0331,0.1581,0.12,LR,0,0.0658,0.1102,0.9986,-0.9834,-0.0611
XGB Classifier,0.0331,0.1581,0.15,LR,0,0.066,0.1132,0.9986,-0.9834,-0.0618
\end{filecontents*}
\DTLloadrawdb[autokeys]{adult}{data/adult.csv}
\DTLmaxforkeys[\DTLiseq{\state}{0}][\state=Column6]{adult}{Column9}{\maxIncrease}
\DTLminforkeys[\DTLiseq{\state}{0}][\state=Column6]{adult}{Column10}{\maxDecrease}
\DTLminforkeys[\DTLiseq{\state}{0}][\state=Column6]{adult}{Column11}{\maxDiff}

\begin{table*}[!b]
\vspace*{1.5in}
\fontsize{9}{10}\selectfont 
\centering
\csvreader[autotabular,
filter=\equal{\state}{0},table head= 
\multicolumn{3}{|c|}{\bfseries Baseline algorithm} & \multicolumn{4}{|c|}{\bfseries Fair algorithm}\\
\hline
     Learner & Loss STD & DP disp & 
     $\varepsilon$ & Learner & Loss STD & DP disp & Max increase & Max decrease & Avg difference \\ 
      \hline
]{data/adult.csv}{unfair learner=\unfairlearner,unfair loss std=\unfairlossstd,unfair DP disp=\unfairDPdisp,unfair DP disp=\unfairDPdisp,fair eps=\faireps,fair learner=\fairlearner,state=\state,fair loss std=\fairlossstd,fair DP disp=\fairDPdisp,max individual increase=\maxindividualincrease,max individual decrease=\maxindividualdecrease,average diff=\averagediff}
{\unfairlearner & \unfairlossstd & \unfairDPdisp & \faireps & \fairlearner & \fairlossstd  & \fairDPdisp & \HighlightCell{\maxindividualincrease}{\maxIncrease} & \HighlightCell{\maxindividualdecrease}{\maxDecrease} & \HighlightCell{\averagediff}{\maxDiff}}%
\caption{Experimental results for Adult data set}
\label{tab:results_adult}
\end{table*}

\begin{filecontents*}{data/communities.csv}
unfair learner,unfair loss std,unfair DP disp,fair eps,fair learner,state,fair loss std,fair DP disp,max individual increase,max individual decrease,average diff
OLS,0.0013,0.4847,0.01,LS,0,0.003,0.0627,0.1246,-0.7217,-0.063
OLS,0.0013,0.4847,0.02,LS,0,0.0028,0.1152,0.1423,-0.7967,-0.0585
OLS,0.0013,0.4847,0.03,LS,0,0.0027,0.1413,0.1246,-0.5467,-0.0563
OLS,0.0013,0.4847,0.04,LS,0,0.0026,0.1595,0.1246,-0.5808,-0.0543
OLS,0.0013,0.4847,0.06,LS,0,0.0023,0.2332,0.1337,-0.5196,-0.0463
OLS,0.0013,0.4847,0.08,LS,0,0.0022,0.2789,0.1246,-0.5129,-0.0423
OLS,0.0013,0.4847,0.1,LS,0,0.0019,0.2881,0.1246,-0.4226,-0.0354
OLS,0.0013,0.4847,0.12,LS,0,0.0018,0.3319,0.1246,-0.3783,-0.0294
OLS,0.0013,0.4847,0.15,LS,0,0.0015,0.3889,0.1246,-0.2717,-0.0211
OLS,0.0013,0.4847,0.17,LS,0,0.0015,0.4126,0.1246,-0.1981,-0.0162
OLS,0.0013,0.4847,0.2,LS,0,0.0014,0.4569,0.1246,-0.1336,-0.0069
OLS,0.0013,0.4847,0.23,LS,0,0.0013,0.4796,0.1246,-0.024,-0.0003
RF Regressor,0.0012,0.4998,0.01,LS,0,0.003,0.0768,0.2383,-0.6314,-0.0657
RF Regressor,0.0012,0.4998,0.02,LS,0,0.0028,0.0872,0.2133,-0.6787,-0.0608
RF Regressor,0.0012,0.4998,0.03,LS,0,0.0027,0.1086,0.1883,-0.6287,-0.0585
RF Regressor,0.0012,0.4998,0.04,LS,0,0.0025,0.1541,0.2633,-0.5537,-0.0556
RF Regressor,0.0012,0.4998,0.06,LS,0,0.0023,0.2049,0.2383,-0.5376,-0.0503
RF Regressor,0.0012,0.4998,0.08,LS,0,0.0021,0.2714,0.2383,-0.4787,-0.0449
RF Regressor,0.0012,0.4998,0.1,LS,0,0.0019,0.3119,0.271,-0.4564,-0.0374
RF Regressor,0.0012,0.4998,0.12,LS,0,0.0018,0.3371,0.296,-0.4814,-0.0334
RF Regressor,0.0012,0.4998,0.15,LS,0,0.0015,0.3773,0.2633,-0.3876,-0.0241
RF Regressor,0.0012,0.4998,0.17,LS,0,0.0015,0.4171,0.321,-0.3314,-0.0187
RF Regressor,0.0012,0.4998,0.2,LS,0,0.0014,0.4561,0.421,-0.2814,-0.0102
RF Regressor,0.0012,0.4998,0.23,LS,0,0.0013,0.4796,0.421,-0.2814,-0.0036
XGB Regressor,0.0014,0.4786,0.01,LS,0,0.003,0.0754,0.2981,-0.5215,-0.0649
XGB Regressor,0.0014,0.4786,0.02,LS,0,0.0028,0.1016,0.3231,-0.4765,-0.0619
XGB Regressor,0.0014,0.4786,0.03,LS,0,0.0029,0.1379,0.2731,-0.57,-0.0594
XGB Regressor,0.0014,0.4786,0.04,LS,0,0.0027,0.1499,0.2731,-0.4682,-0.0571
XGB Regressor,0.0014,0.4786,0.06,LS,0,0.0023,0.1853,0.2731,-0.42,-0.0502
XGB Regressor,0.0014,0.4786,0.08,LS,0,0.0021,0.2688,0.2981,-0.3717,-0.0444
XGB Regressor,0.0014,0.4786,0.1,LS,0,0.0019,0.3109,0.2981,-0.3717,-0.0394
XGB Regressor,0.0014,0.4786,0.12,LS,0,0.0017,0.3119,0.2981,-0.32,-0.0326
XGB Regressor,0.0014,0.4786,0.15,LS,0,0.0015,0.4025,0.2912,-0.2909,-0.0233
XGB Regressor,0.0014,0.4786,0.17,LS,0,0.0015,0.4276,0.3231,-0.2967,-0.0196
XGB Regressor,0.0014,0.4786,0.2,LS,0,0.0014,0.4571,0.3912,-0.2264,-0.0091
XGB Regressor,0.0014,0.4786,0.23,LS,0,0.0013,0.4796,0.4162,-0.2264,-0.0033
\end{filecontents*}
\DTLloadrawdb[autokeys]{communities}{data/communities.csv}
\DTLmaxforkeys[\DTLiseq{\state}{0}][\state=Column6]{communities}{Column9}{\maxIncrease}
\DTLminforkeys[\DTLiseq{\state}{0}][\state=Column6]{communities}{Column10}{\maxDecrease}
\DTLminforkeys[\DTLiseq{\state}{0}][\state=Column6]{communities}{Column11}{\maxDiff}

\begin{table*}[h]
\fontsize{9}{10}\selectfont 
\centering
\csvreader[autotabular,
filter=\equal{\state}{0},table head= 
\multicolumn{3}{|c|}{\bfseries Baseline algorithm} & \multicolumn{4}{|c|}{\bfseries Fair algorithm}\\
\hline
     Learner & Loss STD & DP disp & 
     $\varepsilon$ & Learner & Loss STD & DP disp & Max increase & Max decrease & Avg difference \\ 
      \hline
]{data/communities.csv}{unfair learner=\unfairlearner,unfair loss std=\unfairlossstd,unfair DP disp=\unfairDPdisp,unfair DP disp=\unfairDPdisp,fair eps=\faireps,fair learner=\fairlearner,state=\state,fair loss std=\fairlossstd,fair DP disp=\fairDPdisp,max individual increase=\maxindividualincrease,max individual decrease=\maxindividualdecrease,average diff=\averagediff}
{\unfairlearner & \unfairlossstd & \unfairDPdisp & \faireps & \fairlearner & \fairlossstd  & \fairDPdisp & \HighlightCell{\maxindividualincrease}{\maxIncrease} & \HighlightCell{\maxindividualdecrease}{\maxDecrease} & \HighlightCell{\averagediff}{\maxDiff}}%
\caption{Experimental results for Communities data set}
\label{tab:results_communities}
\end{table*}

\begin{filecontents*}{data/law_school.csv}
unfair learner,unfair loss std,unfair DP disp,fair eps,fair learner,state,fair loss std,fair DP disp,max individual increase,max individual decrease,average diff
OLS,0.0005,0.3919,0.01,LS,0,0.0006,0.1277,0.1234,-0.0581,-0.0017
OLS,0.0005,0.3919,0.02,LS,0,0.0005,0.1753,0.0973,-0.0564,-0.0047
OLS,0.0005,0.3919,0.03,LS,0,0.0006,0.2445,0.1076,-0.0454,-0.0067
OLS,0.0005,0.3919,0.04,LS,0,0.0005,0.3217,0.0936,-0.0473,-0.0083
OLS,0.0005,0.3919,0.06,LS,0,0.0005,0.3833,0.0035,-0.0267,-0.0117
RF Regressor,0.0003,0.3381,0.01,LS,0,0.0006,0.1181,0.1543,-0.2182,-0.0042
RF Regressor,0.0003,0.3381,0.02,LS,0,0.0005,0.2032,0.1333,-0.2267,-0.0069
RF Regressor,0.0003,0.3381,0.03,LS,0,0.0006,0.2254,0.1253,-0.2462,-0.0087
RF Regressor,0.0003,0.3381,0.04,LS,0,0.0005,0.2908,0.1288,-0.2368,-0.0104
RF Regressor,0.0003,0.3381,0.06,LS,0,0.0005,0.3833,0.1253,-0.2462,-0.0139
XGB Regressor,0.0003,0.3797,0.01,LS,0,0.0006,0.1305,0.1301,-0.2239,-0.0051
XGB Regressor,0.0003,0.3797,0.02,LS,0,0.0006,0.1662,0.126,-0.2739,-0.0056
XGB Regressor,0.0003,0.3797,0.03,LS,0,0.0005,0.2212,0.1108,-0.2489,-0.0079
XGB Regressor,0.0003,0.3797,0.04,LS,0,0.0005,0.286,0.1212,-0.2739,-0.01
XGB Regressor,0.0003,0.3797,0.06,LS,0,0.0005,0.3833,0.1051,-0.2739,-0.0134
\end{filecontents*}
\DTLloadrawdb[autokeys]{law_school}{data/law_school.csv}
\DTLmaxforkeys[\DTLiseq{\state}{0}][\state=Column6]{law_school}{Column9}{\maxIncrease}
\DTLminforkeys[\DTLiseq{\state}{0}][\state=Column6]{law_school}{Column10}{\maxDecrease}
\DTLminforkeys[\DTLiseq{\state}{0}][\state=Column6]{law_school}{Column11}{\maxDiff}

\begin{table*}[h]
\fontsize{9}{10}\selectfont 
\centering
\csvreader[autotabular,
filter=\equal{\state}{0},table head= 
\multicolumn{3}{|c|}{\bfseries Baseline algorithm} & \multicolumn{4}{|c|}{\bfseries Fair algorithm}\\
\hline
     Learner & Loss STD & DP disp & 
     $\varepsilon$ & Learner & Loss STD & DP disp & Max increase & Max decrease & Avg difference \\ 
      \hline
]{data/law_school.csv}{unfair learner=\unfairlearner,unfair loss std=\unfairlossstd,unfair DP disp=\unfairDPdisp,unfair DP disp=\unfairDPdisp,fair eps=\faireps,fair learner=\fairlearner,state=\state,fair loss std=\fairlossstd,fair DP disp=\fairDPdisp,max individual increase=\maxindividualincrease,max individual decrease=\maxindividualdecrease,average diff=\averagediff}
{\unfairlearner & \unfairlossstd & \unfairDPdisp & \faireps & \fairlearner & \fairlossstd  & \fairDPdisp & \HighlightCell{\maxindividualincrease}{\maxIncrease} & \HighlightCell{\maxindividualdecrease}{\maxDecrease} & \HighlightCell{\averagediff}{\maxDiff}}%
\caption{Experimental results for Law School data set}
\label{tab:results_law_school}
\end{table*}


\balance
\bibliographystyle{aaai}
\bibliography{sample-base}

\end{document}